\documentclass[]{llncs}

\usepackage[table,xcdraw]{xcolor}
\usepackage{multirow}
\usepackage{svg}
\usepackage{amsmath}
\usepackage{subcaption}

\usepackage{booktabs} % For better-looking tables
\usepackage{tabularx} % For width-adjustable tables
\usepackage{array} % For table column formatting

\definecolor{mygray}{HTML}{EFEFEF}
\usepackage[T1]{fontenc}
\usepackage{graphicx}
\usepackage{hyperref}
\usepackage{color}

\hypersetup{
    colorlinks = true, % Enables the coloring of links
    linkcolor = black, % Color of internal links
    citecolor = black, % Color of citations
    urlcolor = blue, % Color of URLs
}
\begin{document}

\title{Using LLMs for the Extraction and Normalization of Product Attribute Values}

\titlerunning{WDC-PAVE}

\author{Alexander Brinkmann\inst{1}\orcidID{0000-0002-9379-2048} \and
Nick Baumann\inst{1}\orcidID{0009-0001-1215-9153} \and
Christian Bizer\inst{1}\orcidID{0000-0003-2367-0237}}
\authorrunning{A. Brinkmann et al.}

\institute{University of Mannheim, Schloss, 68161 Mannheim, Germany\\
\email{\{alexander.brinkmann@uni-mannheim.de, nick.baumann@students.uni-mannheim.de, christian.bizer@uni-mannheim.de\}}}
\maketitle              % typeset the header of the contribution
\begin{abstract}
Product offers on e-commerce websites often consist of a product title and a textual product description. In order to enable features such as faceted product search or to generate product comparison tables, it is necessary to extract structured attribute-value pairs from the unstructured product titles and descriptions and to normalize the extracted values to a single, unified scale for each attribute. This paper explores the potential of using large language models (LLMs), such as GPT-3.5 and GPT-4, to extract and normalize attribute values from product titles and descriptions. We experiment with different zero-shot and few-shot prompt templates for instructing LLMs to extract and normalize attribute-value pairs. We introduce the Web Data Commons - Product Attribute Value Extraction (WDC-PAVE) benchmark dataset for our experiments. WDC-PAVE consists of product offers from 59 different websites which provide schema.org annotations. The offers belong to five different product categories, each with a specific set of attributes. The dataset provides manually verified attribute-value pairs in two forms: (i) directly extracted values and (ii) normalized attribute values. The normalization of the attribute values requires systems to perform the following types of operations: name expansion, generalization, unit of measurement conversion, and string wrangling.  Our experiments demonstrate that GPT-4 outperforms the PLM-based extraction methods SU-OpenTag, AVEQA, and MAVEQA by 10\%, achieving an F1-score of 91\%. For the extraction and normalization of product attribute values, GPT-4 achieves a similar performance to the extraction scenario, while being particularly strong at string wrangling and name expansion. 

\end{abstract}

\section{Introduction}
\label{sec:introduction}

\begin{figure}[]
\centering
\includegraphics[width=1\textwidth]{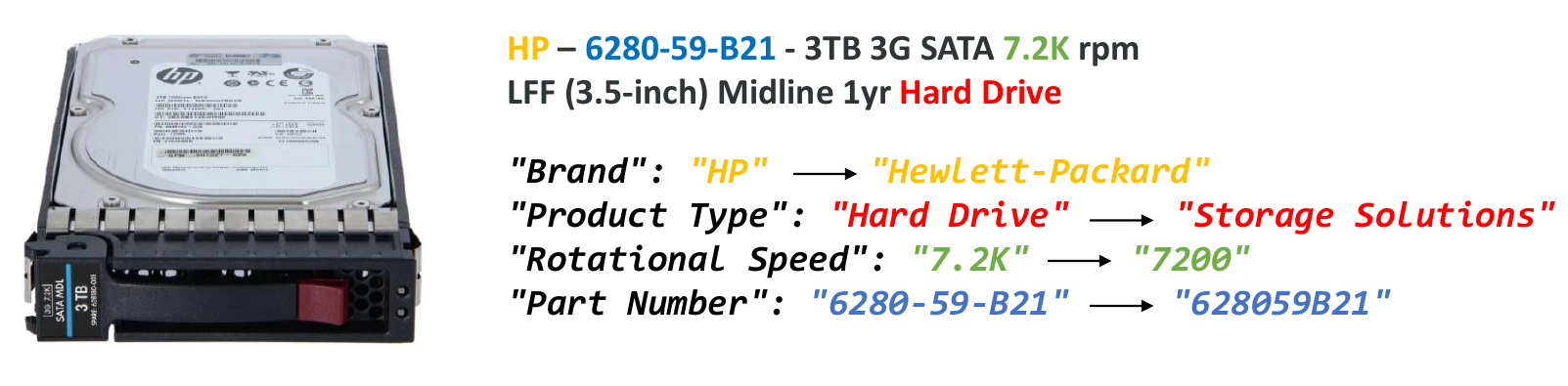}
\caption{Product offer with extracted and normalized attribute-value pairs.} 
\label{fig:example}

\end{figure}

Product attribute value extraction (PAVE) identifies attribute values in product titles and descriptions. After normalizing the extracted attribute values to attribute-specific scales, they are used for tasks such as faceted product search or product comparison.
Figure \ref{fig:example} shows an example of a product offer and attribute-value pairs that have been extracted from the product title. For each attribute, the extracted and the normalized value are displayed. 

Existing methods for PAVE often require large amounts of domain-specific training data to fine-tune pre-trained language models (PLM) that
%learn extraction rules~\cite{ghani2006text,putthividhya2011bootstrapped,wong2009scalable}, 
label attribute value sequences~\cite{jain2021learning,yan2021adatag,zheng2018opentag} or extract attribute values using question answering~\cite{wang2020learning,yang2022mave}. The methods focus on identifying the sequences of tokens that form attribute values but do not cover the normalization of the extracted values.
Motivated by the success of large language models (LLMs) in related NLP tasks~\cite{wei2022emergent} and other information extraction use cases~\cite{agrawal2022large,brinkmann2023product,parekh2023geneva}, this paper explores the potential of LLMs for the following PAVE tasks: (i) direct extraction, (ii) extraction with normalization, and (iii) normalization. The objective of the direct extraction task is to extract sequences of tokens that form attribute values from product titles and descriptions~\cite{wang2020learning,xu2019scaling,yan2021adatag,yang2022mave}. The goal of the extraction with normalization task is to extract and normalize attribute-value pairs in a single step.
The normalization task aims at normalizing attribute values that were extracted by a 
separate, preceding extraction step~\cite{jaimovitch2023can}.
Since current benchmarks for extracting product attribute values
are designed for measuring extraction quality and do not cover the normalization of attribute values~\cite{xu2019scaling,yang2022mave,zhang2022oa}, we introduce a new benchmark dataset, Web Data Commons - Product Attribute Value Extraction (WDC-PAVE).
Unlike existing benchmark datasets, which contain only data from a single source~\cite{xu2019scaling,yang2022mave,zhang2022oa}, WDC-PAVE consists of 565 product offers originating from 59 different websites that use the schema.org vocabulary. In contrast to related work~\cite{xu2019scaling,yang2022mave}, the 4,687 attribute-value pairs in WDC-PAVE have been manually verified and are available in two formats: (i) extracted and (ii) normalized.
To normalize the attribute-values, the systems need to perform name expansion, generalization, unit of measurement conversion, and string wrangling.
In summary, this paper makes the following contributions:
\begin{enumerate}
    \item We propose prompt templates for instructing LLMs to extract and normalize attribute-value pairs from product titles and descriptions. The templates cover use cases with and without training data. In contrast to existing work on attribute value normalization~\cite{jaimovitch2023can}, the templates exploit the attribute value context for the normalization.
    \item We introduce the WDC-PAVE benchmark consisting of 565 heterogeneous product offers and 4,687 manually verified attribute-value pairs. The benchmark supports three tasks:  (i) direct extraction, (ii) extraction with normalization, and (iii) normalization. In contrast to existing benchmarks~\cite{yang2022mave,xu2019scaling,zhang2022oa}, WDC-PAVE covers value extraction and value normalization. 
    \item We experimentally compare the extraction performance of GPT-3.5 and GPT-4 to the PLM-based extraction methods SU-OpenTag~\cite{xu2019scaling}, AVEQA~\cite{wang2020learning}, and MAVEQA~\cite{yang2022mave} on WDC-PAVE. GPT-4 achieves the overall best results with an F1-score of 91\% and outperforms the PLM baselines by 10\%.
    \item We experiment with extracting and normalizing attribute values in a single step using GPT-3.5 and GPT-4. 
    Given 10 example attribute values and 5 demonstrations, GPT-4 again reaches and overall F1-score of 91\%. The model performs particularly well for string wrangling and name expansion with F1-scores of 95\% and 98\% respectively.
    \item We experiment with normalizing previously extracted attribute values using GPT-3.5 and GPT-4. 
    Given 10 example attribute values and 5 demonstrations, GPT-4 reaches a F1-score of 96\%, which is 5\% higher than in the extraction and normalization scenario.
    %This indicates that the extraction is the more challenging for the LLM than the normalization of the values.

\end{enumerate}

The paper is structured as follows:
Section \ref{sec:wdc_product_attribute_value_extraction} introduces the benchmark dataset WDC-PAVE. Section \ref{sec:experimental_setup} describes the experimental setup. Section \ref{sec:extract_only}, Section \ref{sec:extract_and_normalize}, and Section \ref{sec:normalize} discuss the experimental results for the scenarios: (i) direct extraction, (ii) extraction with normalization, and (iii) normalization of product attribute values. Related work in discussed in Section \ref{sec:related_work}.
The WDC-PAVE benchmark and the code for replicating the experiments are available online\footnote{\url{https://github.com/wbsg-uni-mannheim/wdc-pave}}.

\section{The WDC-PAVE Benchmark}
\label{sec:wdc_product_attribute_value_extraction}

This section introduces the WDC-PAVE benchmark. First, we describe the collection of product offers and attribute-value pairs using schema.org\footnote{\url{https://schema.org/}} annotations and product specification tables within web pages. Second, we present profiling statistics about the WDC-PAVE benchmark. Third, we introduce the normalization operations.
%The dataset, which includes product offers, extracted attribute values and normalized attribute values, is available in our code repository.

\vspace{.1cm}\noindent\textbf{Data collection.}
The Web Data Commons (WDC)\footnote{\url{https://webdatacommons.org/}} project extracts structured data from the Common Crawl\footnote{\url{https://commoncrawl.org/}} and provides the extracted data for public download.
The WDC Product Data Corpus (WDC LSPM)\footnote{\url{https://webdatacommons.org/largescaleproductcorpus/v2/}}~\cite{primpeli2019wdc} is one of the extracted datasets. It consists of over 26 million product offers originating from 79 thousand different websites which employ the schema.org vocabulary to annotate structured product data within their HTML pages. The offers are classified into 26 product categories. In addition to schema.org annotations, WDC LSPM extracts attribute-value pairs from specification tables found in the web pages. The attributes in these pairs are product category-specific, such as the number of processor cores of a computer. Category-specific attributes are not part of the schema.org vocabulary and therefore are not explicitly annotated in the web pages. We clean the product offers and attribute-value pairs in the WDC LSPM corpus, omitting those with missing titles, descriptions, or specification tables, and those with descriptions exceeding 1,000 characters. In addition, HTML and language tags are stripped away, and only product offers in English are kept. 
We select the five categories 'Computers and Accessories', 'Jewelry', 'Grocery and Gourmet Food', 'Office Products', and 'Home and Garden' for WDC-PAVE because they contain a large number of product offers and attribute-value pairs after pre-processing. Subsequently, a random sample of product offers is drawn for each category, with the objective to manually verify their attribute-value pairs. Based on the sampled product offers a fixed set of attributes per category is determined. 
% Todo - Done: Im Folgenden ist unklar wie die Werte aus den Spec-Tables und den annotationen zusammenspielen. Warum werden die Spec-Table Werte überprüft wenn doch später andere Surface-Forms im Gold Standard landen? Bitte besser erklären.
As the attribute-value pairs in the specification tables are heterogeneously annotated on the different websites, a human annotator is required to verify that each attribute value is a sub-string of the title or the description, and that it semantically fits the attribute. Additionally, the human annotator adds attribute values that are not contained in the specification tables but are mentioned in the product offer to the gold standard.
%The initial attribute-value pairs are derived from the specification tables of the product offers. These attribute-value pairs are manually verified by a human annotator. Additionally, the human annotator adds attribute values to the gold standard that are mentioned in the title and the description. Each attribute value is a sub-string of the offer's title or description. 
If an attribute is not referenced in the title or the description, the value "n/a" is assigned to this attribute.

\vspace{.1cm}\noindent\textbf{Dataset Statistics.} Table \ref{tab:wdc-pave} shows profiling statistics describing the WDC-PAVE benchmark dataset. The dataset consists of 4,687 attribute-value pairs from 565 product which originate from 59 different websites. 
% Todo - Done: Bitte auch Erklärung überarbeiten. Wollt ihr sagen, dass 49% der Werte fehlen, e.g. "n/a" sind? Wenn ja, dann bitte so hinschreiben.
Overall, 45\% of the attribute-value pairs hold the attribute value "n/a" meaning the attribute is neither mentioned in the title nor in the description.
The dataset contains 2,011 unique attribute values so that each attribute has on average 54 unique values.

\begin{table}[h!]
%\vspace{-5mm}
\centering
\caption{Statistics for WDC-PAVE}
\label{tab:wdc-pave}
\begin{tabular}{l|r|r|r|r|r|r}
\toprule
         & \multicolumn{1}{l|}{Home \&}  & \multicolumn{1}{l|}{Computers \&}  & \multicolumn{1}{l|}{Grocery \&} & \multicolumn{1}{l|}{Office}  &  & \\ 
\textbf{Category}         & \multicolumn{1}{l|}{Garden} & \multicolumn{1}{l|}{Accessories} &  \multicolumn{1}{l|}{Gourmet} &  \multicolumn{1}{l|}{Products} & \multicolumn{1}{l|}{Jewelry} & \multicolumn{1}{l}{Overall}  \\ 
\midrule
\textbf{Unique Attributes}    & 8                                                                             & 11                                                                                       & 5                                                                                  & 10                                                                              & 3  &  37   \\
\textbf{Attribute-Value Pairs}       & 1,136                                                                           & 1,914                                                                                     & 160                                                                                & 1,180                                                                            & 297  &  4,687 \\
\textbf{Unique Values}       & 493                                                                             & 576                                                                                       & 136                                                                                 & 658                                                                              & 148  & 2,011 \\
\textbf{Unique Norm. Values} & 305 & 343 & 92 & 388 & 116 & 1,244 \\
\textbf{Product Offers}           & 142                                                                            & 174                                                                                      & 32                                                                                 & 118                                                                             & 99  &  565 \\
\textbf{Host Websites}       & 16                                                                             & 10                                                                                       & 6                                                                                 & 10                                                                              & 17  & 59 \\
\bottomrule
\end{tabular}
\end{table}

\vspace{.1cm}\noindent\textbf{Data Normalization.}
Each of the 37 attribute in the dataset requires to be normalized before being usable for applications such as faceted product search. 
We have identified four normalization operations. Table \ref{tab2} illustrates the normalization operations with examples for selected attributes. Name Expansion deals with the expansion of abbreviated attribute values such as "HP" into their non-abbreviated form, e.g. "Hewlett-Packard". Generalization assigns attribute values to broader categories, e.g. the color "Neon Lime Green" to the more general category "Green".
Unit of Measurement Normalization converts an attribute value to an attribute-specific target unit of measurement and format, such as the weight value "20-lb." to "9.06", which represents the weight in kilograms (kg).
String Wrangling normalizes attribute values to a specific format by for example replacing words with numbers or removing non-alphanumeric characters, e.g. the value "CTW-4M(208)" would be normalized to "CTW4M208".
Each attribute is assigned to one of the normalization operations.
After normalizing the 2,011 unique values in WDC-PAVE, the dataset contains 1,244 unique normalized attribute values.

\begin{table}[]
\caption{Overview of attribute value normalization operations in WDC-PAVE}
\label{tab2}
\begin{tabular}{l|l|l}
\textbf{Operation}                                                               & \textbf{Attributes}                                                                                                    & \textbf{Examples}                                                                                                                                                                                                 \\ \hline
\textbf{\begin{tabular}[c]{@{}l@{}}Name \\ Expansion\end{tabular}}                      & \begin{tabular}[c]{@{}l@{}}Manufacturer,\\ Generation,\\ Capacity, Cache\end{tabular}                                  & \begin{tabular}[c]{@{}l@{}}"HP" $\rightarrow$ "Hewlett-Packard" \\ "PII" $\rightarrow$ "Pentium II" \\ "G1" $\rightarrow$ "Generation 1"\end{tabular}                                                              \\ \hline
\textbf{Generalization}                                                                 & \begin{tabular}[c]{@{}l@{}}Product Type,\\ Color, Processor \\ Type\end{tabular}                                       & \begin{tabular}[c]{@{}l@{}}"Oatmeal" $\rightarrow$ "Snacks and Breakfast"\\ "Sparkling Juices" $\rightarrow$ "Beverages"\\ "Neon Lime Green" $\rightarrow$ "Green"\end{tabular} \\ \hline
\textbf{\begin{tabular}[c]{@{}l@{}}Unit of \\ Measurement\\ Normalization\end{tabular}} & \begin{tabular}[c]{@{}l@{}}Dimensions,\\ Paper Weight,\\ Size/Weight,\\ Rotational Speed,\\ Pack Quantity\end{tabular} & \begin{tabular}[c]{@{}l@{}}"7"" $\rightarrow$ "17.8"\\ "164 ft" $\rightarrow$ "4998.7"\\ "20-lb." $\rightarrow$ "9.06" (kg)\\ "0.31 oz" $\rightarrow$ "879" (g)\\ "10k" $\rightarrow$ "10000"\end{tabular}        \\ \hline
\textbf{\begin{tabular}[c]{@{}l@{}}String \\ Wrangling\end{tabular}}                    & \begin{tabular}[c]{@{}l@{}}Identifiers, Ports\\ Processor Core,\\ Retail UPC,\\ Brand\end{tabular}                     & \begin{tabular}[c]{@{}l@{}}"CTW-4M(208)" $\rightarrow$ "CTW4M208"\\ "Dual Port" $\rightarrow$ "2"\\ "4-Core" $\rightarrow$ "4"\\ "Quaker Foods" $\rightarrow$ "QUAKER FOODS"\end{tabular}                        
\end{tabular}

\end{table}

\section{Experimental Setup}
\label{sec:experimental_setup}

For our experiments, we split the WDC-PAVE dataset into a training set with 211 product offers and 1,750 attribute-value pairs as well as a test set with 354 product offers and 2,937 attribute-value pairs, stratified by product category. 
%A random subset of 20\% of the training records per product category is used to select in-context demonstrations and example values to simulate  scenarios in which only a small number of labeled attribute-value pairs are available as supervision. Attributes that do not require value normalization are removed from both the training and test sets.
% Todo Alex-Done: Bitte explizit sagen wie groß die verwendeten Trainings und Test sets tatsächlich sind? Wie viele Attributewerte werden benutzt (37 von 70 attributen)? Wie viele Trainingsbeispiele (20% von 75%)?
We access GPT-3.5-turbo-16k-0613 referred to as GPT-3.5 and GPT-4-0613 referred to as GPT-4 through the OpenAI API\footnote{\url{https://platform.openai.com/docs/api-reference}}. The temperature parameter of the LLMs is set to 0 to reduce the randomness. GPT-3.5 and GPT-4 are not fine-tuned. Instead, we select semantically similar demonstration product offers for in-context learning.
As baselines for the experiments, we fine-tune the PLM-based extraction methods SU-OpenTag~\cite{xu2019scaling}, AVEQA~\cite{wang2020learning} and MAVEQA~\cite{yang2022mave} on the training set. The fine-tuning is executed on a single NVIDIA RTX A6000 GPU. Since, the prompt templates for the LLMs and the PLM-based extraction methods utilize the same training set for demonstration selection and fine-tuning, respectively, we assume that this is a fair comparison.
% Todo Alex - Done: Bitte klar sagen, dass ihr die PLM-nur mit 20% des Trainingsets trainiert, falls ihr dass so macht => wäre ein  eher merkwürdiges Setup. Oder sagen, dass ihr das ganze Trainingset benutzt (nur 37 attribute?). Stellt ihr die tatsächlihc genutzen Datasets und das gasamte Dataset beide zum Download bereit?
%Kommentar Alex: Beide Trainingsets sind online verfügbar. Da im Text nur noch von einem Trainingsset die Rede ist, gehe ich auf diesen Punkt nicht nochmal ein.
For the evaluation, we follow related work and calculate F1-scores based on the exact match between the predicted and the ground truth attribute values~\cite{brinkmann2023product,wang2020learning,xu2019scaling,yan2021adatag,yang2022mave}.

\section{Direct Extraction}
\label{sec:extract_only}

\begin{figure}[h!]
\centering
\includegraphics[width=1\textwidth]{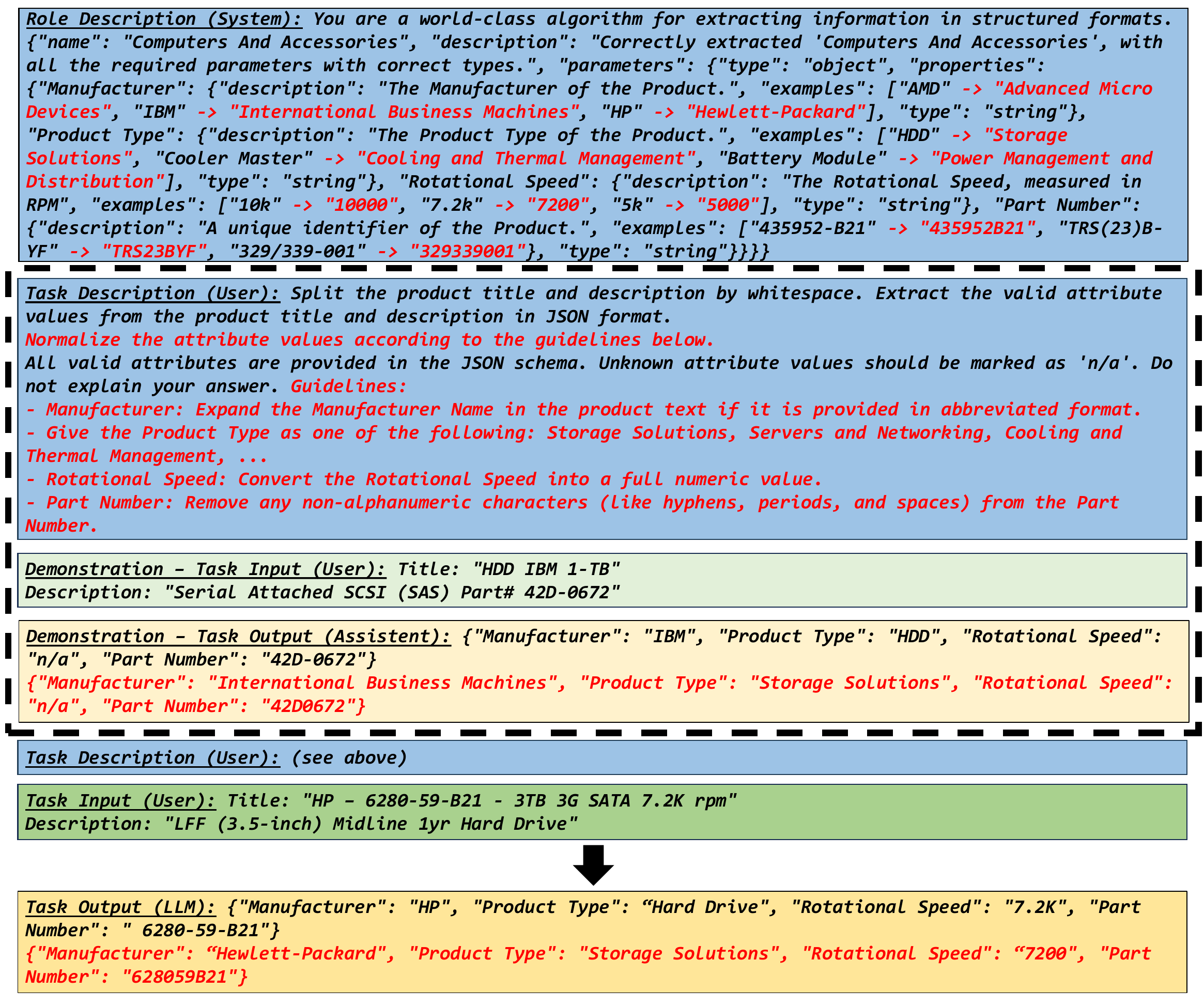}
\caption{Example prompt. The parts set in back font are used for the extraction. The red parts are added for the extraction with normalization task).} 
\label{fig:prompt_template}
\end{figure}

This section compares various prompt templates for extracting attribute-value pairs from the product offers in WDC-PAVE. Following Brinkmann et al.~\cite{brinkmann2023product}, the prompts ask the LLM to extract all attributes of the target schema in a single step. Figure \ref{fig:prompt_template} shows an example of a complete prompt. 

\vspace{.1cm}\noindent\textbf{Prompt Templates.} Each prompt template consists of up to six building blocks visualized by different background colors.
The building blocks are role description (\textcolor{blue}{blue}), task description (\textcolor{blue}{blue}), task input (\textcolor{green}{green}), 
 and task output (\textcolor{orange}{orange}), demonstration input (\textcolor{green}{green}) and demonstration output (\textcolor{yellow}{yellow}).
The role description is a system message that defines the overall goal of the LLM as well as the target schema for the extraction including attribute descriptions and example values. The target schema is encoded using JSON-schema\footnote{\url{https://json-schema.org/}} as this representation proved to be most effective in related work~\cite{brinkmann2023product}. The task description is a user message that provides instructions for the attribute-value extraction. The task input is a user message containing the product title and description. The task output contains the LLM's response with the extracted attribute-value pairs in JSON format. The input and output of the demonstrations are user and assistant messages containing product offers that are semantically similar to the target product offer in the task input. To calculate the semantic similarity, the target product offer and the training demonstrations are embedded using OpenAI's embedding model \texttt{text-embedding-ada-002}\footnote{\url{https://platform.openai.com/docs/guides/embeddings/}}. The training demonstrations with the highest cosine similarity to the product offer are added into the prompt.
% Todo Alex-Done: Fill in metric. Be explicit about which training set? Full or 20%?
Figure \ref{fig:prompt_template} contains instructions for extraction and normalization. The red text is added to the prompts for the extraction with normalization task and the normalization task, which are discussed in Section \ref{sec:extract_and_normalize} and Section \ref{sec:normalize}.

\vspace{.1cm}\noindent\textbf{Discussion of Results.} We assess the impact of zero-shot and few-shot prompt template configurations on GPT-3.5 and GPT-4 performance. Table \ref{tab:extraction_results} shows the F1-scores, the average number of tokens per prompt, and the cost in \$ per 1,000 extracted attribute values of the prompt template configurations with different amounts of example values (Val.) and the combination of 10 example values and different amounts of demonstrations (Dem.). The highest F1-score with and without demonstrations is marked in bold. For the cost calculation, we use the OpenAI prices as of April 2024\footnote{\url{https://openai.com/pricing}}. The results show that GPT-3.5 and GPT-4 benefit from ten example values, achieving F1-scores of around 80\%. Demonstrations further improve the performance of GPT-3.5 and GPT-4. Ten demonstrations allow GPT-4 to achieve the highest F1-score of 91\%.
Both, example values and demonstrations implicitly guide the LLMs to extract exactly the surface form of the attribute values that is used in the offer and is expected by the ground truth.
The usage of example values and demonstrations significantly increases the length and the cost of the prompts. Adding three demonstrations increases the cost of extracting 1,000 attribute values with GPT-4 by 1\$. In order to enhance GPT-4's performance by an additional 1.6\%, an additional 2\$ per 1,000 extracted attribute values must be spent. 

\begin{table}[]
\centering
\caption{Results for the direct extraction task.}
\label{tab:extraction_results}
\begin{tabular}{@{}cc|cc|cc|cc@{}}
\toprule
     &      & \multicolumn{2}{c|}{F1}                                 & \multicolumn{2}{c|}{Average length}           & \multicolumn{2}{c}{\$ per 1K values} \\ 
Val. & Dem. & GPT-3.5 & GPT-4 & GPT-3.5 & GPT-4 & GPT-3.5    & GPT-4    \\ \midrule
0                        & 0                        & 70.66                       & 74.40                     & 750                         & 745                       & 0.09                         & 0.10                       \\
3                        & 0                        & 77.11                       & 78.77                     & 985                         & 973                       & 0.12                         & 1.25                       \\
5                        & 0                        & 78.51                       & 77.52                     & 1097                        & 1114                      & 0.13                         & 1.40                       \\
10                       & 0                        & 80.54                       & 79.65                     & 1334                        & 1338                      & 0.15                         & 1.66                       \\ \midrule
10                       & 3                        & 86.91                       & 88.94                     & 2274                        & 2205                      & 0.26                         & 2.61                       \\
10                       & 5                        & 86.93                       & 88.15                     & 2776                        & 2735                      & 0.32                         & 3.21                       \\
10                       & 10                       & \textbf{88.02}                       & \textbf{90.54}                     & 3975                        & 3974                      & 0.43                         & 4.60                       \\ \bottomrule
\end{tabular}
\end{table}

\vspace{.1cm}\noindent\textbf{Comparison of LLMs and PLMs.}
We now compare the performance of GPT-3.5 and GPT-4 with the prompt template that uses ten example values and ten demonstrations to the PLM baselines SU-OpenTag~\cite{xu2019scaling}, AVEQA~\cite{wang2020learning} and MAVEQA~\cite{yang2022mave}. The baselines are fine-tuned on the same training set used to select example values and task demonstrations for the LLM prompt templates. The results in Table \ref{tab:comparison_to_PLMs} show that GPT-4 outperforms the best PLM baseline AVEQA by 10\% F1.
It is important to mention that the training set with 1,750 attribute-value pairs is small. The training sets of the existing benchmark datasets MAVE~\cite{yang2022mave} and AE-110k~\cite{xu2019scaling} contain 3.8 million and 84.7 thousand attribute-value pairs. As shown by Brinkmann et al.~\cite{brinkmann2023product}, it can be expected that with additional training data the performance of the PLM baselines increases whereas the performance of the LLMs only marginally improves.
% Todo Alex-Done: Nachschauen ob die Größe des Trainingssets (20% von 75%) mit der Größe von Trainignsset in der PLM-related Work vergleichbar ist. Falls nein, explizit sagen, dass wir die PLMs mit wenig Daten trainieren bzw. zusätzlich auch vergleichen zu PLMs mit vollem Trainingset. 

\begin{table}[]
\centering
\caption{Comparison of LLM- and PLM-based methods.}
\label{tab:comparison_to_PLMs}
\begin{tabular}{@{}l|cc|ccc@{}}
\toprule
               & GPT-4 & GPT-3.5  & AVEQA & MAVEQA & SU-OpenTag \\ \midrule
F1             & \textbf{90.54} & 88.02                                           & 80.83                     & 65.10  & 60.44      \\
$\Delta_1$ to GPT-4 & - & -2.52                                             & -9.71                    & -25.44 & -30.10     \\ \bottomrule
\end{tabular}
\end{table}
\section{Extraction with Normalization}
\label{sec:extract_and_normalize}

This section evaluates prompt templates instructing LLMs to extract and normalize attribute values in a single step. 
To instruct the LLM on how to normalize attribute values, the prompt template in Figure \ref{fig:prompt_template} is extended with the texts set in red font.
The extensions add normalization guidelines to the task description and mappings between example values and their normalized forms to the target schema. The attribute values in the demonstrations are normalized.

\vspace{.1cm}\noindent\textbf{Discussion of Results.} As in Section \ref{sec:extract_only}, we measure how adding zero, three, five and ten example values without demonstrations and ten example values with three, five and ten demonstrations to the prompt affects the performance of GPT-3.5 and GPT-4.
Table \ref{tab:extraction_and_normalization} shows the results of the experiments. The best F1-scores per model are marked in bold. Without demonstrations, GPT-4 benefits from normalized example values and reaches an F1-score of 86\%, which is 12\% better than the zero-shot configuration. In contrast to the extraction task, example values only marginally improve the extraction and normalization performance of GPT-3.5. However, adding five demonstrations improves the performance of both LLMs, with GPT-4 achieving an F1-score of 91\% and outperforming GPT-3.5 by 5\%. The prompts for the extraction with normalization task are longer than the prompts for the extraction task because of the additional normalization instructions. At the same time, the best performance is achieved with 5 instead of 10 demonstrations, resulting in a cost reduction of 1\$ per 1,000 extracted attribute values for GPT-4 (see rightmost column in Table \ref{tab:extraction_and_normalization}).

%\begin{table}[]
%\centering
%\caption{F1-scores for the extraction with normalization scenario.} 
%\label{tab:extraction_and_normalization}
%\begin{tabular}{l|r|rrr|rrr}
%\hline
%        & \multicolumn{4}{c|}{}                                                                             & \multicolumn{3}{c}{10 Example Values \&}                                                     \\ %\cline{1-8}  
%\textbf{LLM}   & 0 Val. & 3 Val. & 5 Val. & 10 Val.         & 3 Dem. & 5 Dem. & 10 Dem. \\  \hline
%\textbf{GPT-3.5} & 68.86                          & 69.75                          & 69.91                 & \multicolumn{1}{l|}{69.36}         & 85.90                     & 86.33            & 86.81                      \\
%\textbf{GPT-4}   & 74.19                             & 83.77                          & 84.90                           & \multicolumn{1}{l|}{\textbf{85.60}} & 91.18                     & \textbf{91.32}                     & 91.31             \\ \hline
%\end{tabular}
%\vspace{-5mm}
%\end{table}

\begin{table}[]

\centering
\caption{Results for the extraction with normalization task.}
\label{tab:extraction_and_normalization}
\begin{tabular}{@{}cc|rr|rr|rr@{}}
\toprule
     &      & \multicolumn{2}{c|}{F1-score}                                 & \multicolumn{2}{l|}{Avg. Tok. per Prompt}           & \multicolumn{2}{l}{\$ per 1k Attr. Val.} \\ 
Val. & Dem. & GPT-3.5 & GPT-4 & GPT-3.5 & GPT-4 & GPT-3.5    & GPT-4    \\ \midrule
0                        & 0                        & 68.62                       & 74.19                     & 902                         & 889                      & 0.1043                         & 1.0820                       \\
3                        & 0                        & 71.32                       & 83.77                     & 1156                        & 1155                      & 0.1327                         & 1.3774                       \\
5                        & 0                        & 70.17                       & 84.90                     & 1316                        & 1312                      & 0.1505                         & 1.5531                       \\
10                       & 0                        & 68.82                       & 85.60                     & 1715                        & 1673                      & 0.1954                         & 1.9578                       \\ \midrule
10                       & 3                        & 85.68                       & 91.18                     & 2708                        & 2702                      & 0.3057                         & 3.1028                       \\
10                       & 5                        & \textbf{86.37}                       & \textbf{91.32}                     & 3040                        & 3079                      & 0.3428                         & 3.5247                       \\
10                       & 10                       & 86.30                       & 91.31                     & 4024                        & 4031                      & 0.4529                         & 4.5901      \\ \bottomrule                
\end{tabular}
\end{table}

\vspace{.1cm}\noindent\textbf{Analysis of Normalization Operations.} We now analyze the performance of the LLMs on the normalization operations in detail. Table \ref{tab:normalization_operations} shows
%Todo Alex - Done: What do you mean with configuration here?
 the F1-scores per normalization operation for the prompt templates zero-shot , with ten example values and with ten example values and five demonstrations. Previous research has shown that GPT-3.5 performance is weaker on tasks requiring reasoning or calculation than on tasks involving manipulation of free text or names~\cite{jaimovitch2023can}. Our results support these observations. GPT-3.5 and GPT-4 are particularly strong at name expansion and string wrangling. GPT-4 achieves F1-scores of 98\% and 97\%. Unit of measurement conversion requires calculations and is the most challenging operation for GPT-3.5 and GPT-4. Example values and demonstrations improve GPT-4's zero-shot performance by 22\%, leading to an F1-score of 83.5\%. 
Compared to the extraction task, we observe that generalizing attribute values simplifies the task for GPT-4, possibly because it can use its background knowledge to generalize the values.
For attributes like 'Product Type', we observe that GPT-4 benefits from the generalization by an average of 7\% across all product categories if example values and demonstrations from a training set are provided.

%Todo Alex - Done: Remove bold font for operation and model names. 
\begin{table}[t]
\caption{F1-scores by normalization operation for the extraction with normalization task.}
\label{tab:normalization_operations}
\centering
\begin{tabular}{l|rrr|rrr}
\hline
 & \multicolumn{3}{c|}{\textbf{GPT-3.5}}                                                                                                 & \multicolumn{3}{c}{\textbf{GPT-4}}                                                                                                \\ \cline{2-7}  
\begin{tabular}[c]{@{}c@{}}\\ \textbf{Normalization Operation}\end{tabular}    & 0 Val. & 10 Val. & \begin{tabular}[c]{@{}c@{}}10 Val. \\ 5 Dem.\end{tabular} & 0 Val. & 10 Val. & \begin{tabular}[c]{@{}c@{}}10 Val. \\ 5 Dem.\end{tabular} \\ \hline
Name Expansion & 41.61                                                   & 42.15                                             & \textbf{94.50}                                    & 48.64                                                   & 93.60                                              & \textbf{98.27}                                   \\
Generalization                          & 75.27                                                   & 76.63                                               & \textbf{82.16}                                   & 76.48                                                   & 79.82                                              & \textbf{88.56}                                    \\
Unit of Measurement Norm.        & 51.16                                                   & 47.64                                             & \textbf{76.24}                                    & 61.76                                                   & 73.86                                               & \textbf{83.50}                                   \\
String Wrangling                        & 87.07                                                   & 83.78                                              & \textbf{93.37}                                    & 92.41                                                   & \textbf{97.37}                                              & 95.19                                    \\ \hline
\end{tabular}
\end{table}

\section{Normalization}
\label{sec:normalize}

This section compares different prompt templates instructing LLMs to normalize attribute values which have been extracted by a separate preceding extraction step. 
As in Section \ref{sec:extract_and_normalize}, the prompt template in Figure \ref{fig:prompt_template} is extended by the text set in red font. The attribute values to be normalized are added to the task input block in addition to the product title and the description, which can be exploited as context in the normalization process. 

\vspace{.1cm}\noindent\textbf{Discussion of Results.} Like in Section \ref{sec:extract_only} and Section \ref{sec:extract_and_normalize}, we assess how zero, three, five and ten example values without demonstrations and ten example values with three, five and ten demonstrations selected from the training set impact the performance of GPT-3.5 and GPT-4.
Table \ref{tab:normalization} shows the results of the experiments. The highest F1-scores are marked in bold. Similar as in the previous tasks, we observe that both example values and demonstrations improve the F1-scores of GPT-3.5 and GPT-4. The effect of adding demonstrations for GPT-3.5 is marginal while GPT-4 gains 4\% F1-score.

\begin{table}[]
\centering
\caption{Results for the normalization task.}
\label{tab:normalization}
\begin{tabular}{@{}cc|rr|rr|rr@{}}
\toprule
     &      & \multicolumn{2}{c|}{F1}                                 & \multicolumn{2}{c|}{Average length}           & \multicolumn{2}{c}{\$ per 1K values} \\ 
Val. & Dem. & GPT-3.5 & GPT-4 & GPT-3.5 & GPT-4 & GPT-3.5    & GPT-4    \\ \midrule
0                        & 0                        & 82.81       & 86.41    & 974                         & 974                       & 0.1120                         & 1.1813                       \\
3                        & 0                        & 89.99       & 92.00    & 1242                        & 1260                      & 0.1420                         & 1.5015                       \\
5                        & 0                        & 91.16       & 92.11    & 1378                        & 1376                      & 0.1573                         & 1.6318                       \\
10                       & 0                        & 90.61       & 92.44    & 1732                        & 1732                      & 0.1969                         & 2.0292                       \\ \midrule
10                       & 3                        & 91.68       & 95.76    & 2941                        & 2961                      & 0.3321                         & 3.3931                       \\
10                       & 5                        & \textbf{90.98}       & 96.06    & 3486                        & 3548                      & 0.3931                         & 4.0495                       \\
10                       & 10                       & 90.94       & \textbf{96.21}    & 4795                        & 4811                      & 0.5395                         & 5.4624                       \\ \bottomrule
\end{tabular}
\end{table}

\vspace{.1cm}\noindent\textbf{Analysis of Normalization Operations.} We now analyze the normalization operations in detail. Table \ref{tab:normalization_operations} shows the F1-scores per normalization operation for the prompt templates zero-shot, with ten example values and with ten example values and five demonstrations.
%Todo Alex: What best configuration? The table contains various results for different prompts.
The results show that string wrangling can be well handled by LLMs if the attribute values have already been extracted. The other normalization operations require example values and demonstrations to reach high F1-scores.  Compared to the extraction with normalization task, the unit of measurement conversion results for GPT-3.5 and GPT-4 improve by 17\% and 14\% if the values have already been extracted. This shows how challenging the combination of extraction and normalization for the LLMs is given that unit of measurement conversions need to be performed. In contrast, the generalization operation remains as challenging as in the extraction with normalization task.

\begin{table}[]
\caption{F1-scores by normalization operation for the normalization task.}
\label{tab:normalization_operations}
\centering
\begin{tabular}{l|rrr|rrr}
\hline
 & \multicolumn{3}{c|}{GPT-3.5}                                                                                                 & \multicolumn{3}{c}{GPT-4}                                                                                                \\ \cline{2-7}  
\begin{tabular}[c]{@{}c@{}}\\ Normalization Operation\end{tabular}    & 0 Val. & 10 Val. & \begin{tabular}[c]{@{}c@{}}10 Val. \\ 5 Dem.\end{tabular} & 0 Val. & 10 Val. & \begin{tabular}[c]{@{}c@{}}10 Val. \\ 5 Dem.\end{tabular} \\ \hline
Name Expansion            & 64.47  & 95.18   & \textbf{97.15}           & 84.10  & \textbf{100.00}   & 99.12           \\
Generalization            & 73.58  & \textbf{80.67}   & 79.81           & 78.19  & 82.04   & \textbf{90.72}           \\
Unit of Measurement Norm. & 84.89  & \textbf{93.28}   & 93.15           & 90.28  & 94.07   & \textbf{97.89}           \\
String Wrangling          & 97.30  & 96.71   & \textbf{97.70}           & 93.68  & 98.81   & \textbf{99.32}          \\
 \hline
\end{tabular}
\end{table}

\vspace{.1cm}\noindent\textbf{Comparison across Tasks.} We now analyze how the F1-scores change between (i) direct extraction, (ii) extraction with normalization and (iii) normalization. Table \ref{tab:comparison_of_scenarios} shows the F1-scores of GPT-3.5 and GPT-4 for the zero-shot scenario and the few-shot scenario with ten example values and five demonstrations. In addition, the deltas between the extraction task and the extraction with normalization task ($\Delta_1$) and the delta between the extraction task and the normalization task ($\Delta_2$) are shown. The results indicate that extraction is more challenging for LLMs than normalization. Zero-shot both GPT-3.5 and GPT-4 achieve 12\% higher F1-scores if no extraction is required.
With example values and demonstrations, GPT-3.5 and GPT-4 reach F1-scores that are 4\% and 8\% higher if no extraction is required.

\begin{table}[]
\caption{Comparison of F1-scores across tasks.}
\label{tab:comparison_of_scenarios}
\centering
\begin{tabular}{@{}cc|r|r|rr|rr@{}}
\toprule
 &  &    & & \multicolumn{2}{l|}Extract \&    &  &  \\ 
Val. & Dem. & Model   & Extract &  Normalize & $\Delta_1$(Extr.) & Normalize & $\Delta_2$(Extr.) \\ \midrule
0                       & 0                         & GPT-3.5 & 70,66                          & 68,62                                           & -2,04                                & 82,81                             & +12,15                                 \\
0                       & 0                         & GPT-4   & 74,40                          & 74,19                                           & -0,21                                & 86,41                             & +12,01                                 \\ \midrule
10                      & 5                         & GPT-3.5 & 86,93                          & 86,37                                           & -0,56                                & 90,98                             & +4,05                                  \\
10                      & 5                         & GPT-4   & 88,15                          & 91,32                                           & 3,17                                 & 96,06                             & +7,91                                  \\ \bottomrule
\end{tabular}
\end{table}
\section{Related Work}
\label{sec:related_work}

\vspace{.1cm}\noindent\textbf{Product Attribute Value Extraction.}
Early research on PAVE used domain-specific rules to extract attribute-value pairs~\cite{vandic2012faceted,zhang2009framework,nederstigt2014floppies} from product descriptions.
The initial learning-based methods required extensive feature engineering and did not generalize to unknown attributes and values~\cite{ghani2006text,putthividhya2011bootstrapped}.
Recent works have adopted BiLSTM-CRF architectures~\cite{kozareva2016recognizing,zheng2018opentag} to tag attribute values in product titles.
%OpenTag trains a BiLSTM-CRF model with active learning. 
SU-OpenTag~\cite{xu2019scaling} builds upon OpenTag~\cite{zheng2018opentag} by encoding both a target attribute and the product title using the PLM BERT~\cite{devlin2019bert}. %AdaTag~\cite{yan2021adatag} uses BERT~\cite{devlin2019bert} together with a mixture-of-experts module for attribute value extraction. %TXtract~\cite{karamanolakis2020txtract} integrates a product taxonomy into the extraction model.
\cite{sabeh2022cave,shinzato2022simple,wang2020learning,yang2022mave} approach PAVE as a question-answering task, using different PLMs to encode target attribute, category, and title.
The PLM-based methods SU-OpenTag~\cite{xu2019scaling}, AVEQA~\cite{wang2020learning}, and MAVEQA~\cite{yang2022mave} serve as baselines for the direct extraction task.
%OA-Mine~\cite{zhang2022oa} employs BERT~\cite{devlin2019bert} to mine for unknown attributes and values.
% Commented out in order to shorten paper: Other works deal with attribute value extraction from multiple modalities like texts and images~\cite{iv2017multimodal,wang2022smartave,zhang2023pay,zhu2020multimodal,zou2024implicitave}.

\vspace{.1cm}\noindent\textbf{LLMs for Attribute Value Extraction.}
%LLMs often show better zero-shot performance compared to PLMs and are more robust to unseen examples~\cite{brown2020language} because they are pre-trained on large amounts of text, and have emergent effects due to their model size~\cite{wei2022emergent}.
LLMs have successfully been used for information extraction in various domains~\cite{agrawal2022large,goel2023llms,parekh2023geneva}.
In the context of PAVE recent works experiment with different prompt designs for PAVE using LLMs~\cite{brinkmann2023product,fang2024llm-ensemble}.
We use the prompt templates from~\cite{brinkmann2023product} as a role model for our prompt templates.
In contrast to related work ~\cite{blume2023generative,roy2024exploring,yang2023mixpave}, we do not fine-tune the LLMs, but instead rely on in-context learning via demonstrations and example values.
Early work on attribute value extraction and normalization applied domain-specific normalization rules~\cite{van_rooij2016data,valstar2021apfa}.
Jaimovitch‑López et al.~\cite{jaimovitch2023can} experiment with using GPT-3.5 for attribute value normalization but present the values to be normalized without any context to the model. In contrast, we include the original titles and descriptions into the prompts in order to provide context for the normalization. 

\vspace{.1cm}\noindent\textbf{Benchmarks for Attribute Value Extraction and Normalization.} 
The benchmarks MAVE~\cite{yang2022mave} and AE-110k~\cite{xu2019scaling} are widely used to evaluate methods for PAVE. MAVE relies on an ensemble of five fine-tuned PLMs for determining ground truth annotates. AE-110k~\cite{xu2019scaling} uses values from product specification tables as ground truth. In contrast to these benchmarks, all attribute-value annotations in WDC-PAVE are manually verified. 
To our knowledge, OA-Mine~\cite{zhang2022oa} is the only other publicly available benchmark offering human-verified attribute-value annotations. MAVE, AE-110k, and OA-Mine address value extraction and do not consider value normalization. WDC-PAVE covers both tasks. \cite{jaimovitch2023can} propose an attribute value normalization benchmark including operations such as transforming dates, units of measurement, or names. Unlike WDC-PAVE, their benchmark presents the values to be normalized without any context that can be exploited by the methods.

\section{Conclusion}
\label{sec:conclusion}

This paper investigated the ability of GPT-3.5 and GPT-4 to extract and normalize product attribute values from product offers.
We experimented with different prompt templates that use example values and demonstrations for in-context learning. We introduced the WDC-PAVE benchmark, which features manually verified ground truth values for attribute value extraction as well as value normalization. GPT-4 achieves the best F1-score of 91\% in the extraction task, surpassing the best PLM baseline by 10\%, and shows similar performance for the extraction with normalization task.
% Besser weglassen, da diese Schlussfolgerungen nicht zu unseren Ergbenissen der reinen Normalization task passen: GPT-4 excels in wrangling strings and expanding names, where it can utilize its background knowledge. Normalization tasks that require arithmetic reasoning are challenging for GPT-3.5 and GPT-4.
%A comparison of the tasks reveals that extraction is more challenging than normalization for the LLMs.
A compelling avenue for future research is to give LLMs access to scale-specific functions that the model can decide to invoke for normalizing values.

\newpage
\bibliographystyle{splncs04}
\bibliography{My_Collection_pretty}

\end{document}